\documentclass{article}
\pdfoutput=1

\usepackage[preprint,nonatbib]{neurips_2022}

\usepackage{titletoc}
\usepackage[utf8]{inputenc} 
\usepackage[T1]{fontenc}    
\usepackage{url}            
\usepackage{booktabs}       
\usepackage{amsfonts}       
\usepackage{nicefrac}       
\usepackage{microtype}      
\usepackage{xcolor}         
\usepackage{graphicx}
\usepackage{tikz}
\usepackage{amsmath}
\usepackage{xspace}
\usepackage{wrapfig}

\usepackage{enumitem}
\usepackage{amsmath}
\usepackage{amssymb}
\usepackage{amsthm}
\usepackage[ruled,vlined,linesnumbered]{algorithm2e}

\usepackage[page,header]{appendix}
\usepackage{array}
\usepackage{hyperref}       

\title{Challenges to Solving Combinatorially Hard \\ Long-Horizon Deep RL Tasks}

\author{%
  Andrew C. Li \\
  University of Toronto\\
  Vector Institute \\
  \texttt{andrewli@cs.toronto.edu} \\
  \And
  Pashootan Vaezipoor \\
  University of Toronto \\
  Vector Institute \\
  \texttt{pashootan@cs.toronto.edu} \\
  \AND
  Rodrigo Toro Icarte \\
  Pontificia Universidad Católica de Chile \\
  Centro Nacional de Inteligencia Artificial \\ 
  Vector Institute \\
  \texttt{rodrigo.toro@ing.puc.cl} \\
  \And
  Sheila A. McIlraith \\
  University of Toronto \\
  Vector Institute \\
  Schwartz Reisman Institute \\
  \texttt{sheila@cs.toronto.edu} \\
}
\begin{document}
\definecolor{light-grey}{RGB}{120, 120, 120}
\definecolor{BrickRed}{RGB}{140, 0, 0}
\definecolor{LighterBlue}{RGB}{47, 140, 255}

\newcommand{\cut}[1]{}
\newcommand{\maybecut}[1]{\textcolor{orange}{#1}}
\newcommand{\debating}[1]{\textcolor{red}{#1}}
\newcommand{\added}[1]{\textcolor{red}{#1}}
\newcommand{\donepromised}[1]{{}}
\newcommand{\toresolve}[1]{\textcolor{red}{#1}}
\newcommand{\remove}[1]{\textcolor{green}{#1}}
\newcommand{\removehide}[1]{}
\newcommand{\alt}[1]{\textcolor{brown}{#1}}
\newcommand{\old}[1]{\textcolor{red}{\sout{#1}}}
\newcommand{\new}[1]{\textcolor{BrickRed}{\uline{#1}}}
\newcommand{\todo}[1]{\textcolor{cyan}{({\bf TODO:} #1)}}

\newcommand{\commentsm}[1]{\textcolor{BrickRed}{({\bf SM:} #1)}}
\newcommand{\commentpv}[1]{\textcolor{LighterBlue}{({\bf PV:} #1)}}
\newcommand{\commental}[1]{\textcolor{magenta}{({\bf AL:} #1)}}
\newcommand{\commentrt}[1]{\textcolor{orange}{({\bf RT:} #1)}}



\newcommand{\ppo}[1]{$\text{PPO}^{(\gamma=#1)}$}
\newcommand{\vd}[1]{$\text{PPO}_{\mathcal{VD}}^{(\gamma=#1)}$}
\newcommand{\vdn}{$\text{PPO}_{\mathcal{VD}}$\xspace}



\makeatletter
\def\blfootnote{\gdef\@thefnmark{}\@footnotetext}
\makeatother

\maketitle
\blfootnote{Code is available at \url{https://github.com/andrewli77/combinatorial-rl-tasks}.}

\begin{abstract}

  Deep reinforcement learning has shown promise in discrete domains requiring complex reasoning, including games such as Chess, Go, and Hanabi. However, this type of reasoning is less often observed in long-horizon, continuous domains with high-dimensional observations, where instead RL research has predominantly focused on problems with simple high-level structure (e.g. opening a drawer or moving a robot as fast as possible). Inspired by combinatorially hard optimization problems, we propose a set of robotics tasks 
  which admit many distinct solutions at the high-level, but require reasoning about states and rewards thousands of steps into the future for the best performance.
  Critically, while RL has traditionally suffered on complex, long-horizon tasks due to sparse rewards, our tasks are carefully designed to be solvable without specialized exploration. Nevertheless, our investigation finds that standard RL methods often neglect long-term effects due to discounting, while general-purpose hierarchical RL approaches struggle unless additional abstract domain knowledge can be exploited.
\end{abstract}
\section{Introduction}

Reinforcement learning (RL) is a powerful framework for training sequential decision-making agents when the optimal behaviour is unknown or difficult to specify. Recently, RL has seen impressive strides across a number of domains requiring complex reasoning (e.g. game playing \cite{alphazero, pmlr-v97-foerster19a, anthony2017thinking}, combinatorial optimization \cite{mazyavkina2021reinforcement, kool2018attention}) as well as domains with long horizons and high-dimensional observations \cite{gupta2020relay, duan2016benchmarking, tassa2018deepmind}. Despite this, tasks which require a combination of high-level, temporally abstracted reasoning and low-level control over thousands of timesteps (encompassing many real-world problems) still remain a significant challenge for RL \cite{mirza2020physically}. Often, this failure is due to reward sparsity --- as tasks become increasingly long and difficult to solve, the reward signal also tends to diminish, making it impractical to learn from initially random behaviour.

Drawing inspiration from NP-hard combinatorial optimization problems, we contribute a suite of carefully designed tasks that can be decomposed to yield dense, shaped rewards with the following property: these tasks admit a vast number of feasible high-level solutions but solving them optimally requires the agent to reason over many distinct possible outcomes. We explore the prospect of solving such tasks with deep RL, and our contributions can be summarized as follows:




\begin{itemize}[leftmargin=5.5mm]
    \item Our investigation finds critical issues in applying deep RL to combinatorial RL tasks, even when sparse rewards are not problematic. These issues relate to the discount factor --- undiscounted training is known to be unstable, while we show that discounted RL often leads to myopic behaviour inconsistent with maximizing long-term rewards. 
    \item We propose a simple modification to PPO \cite{schulman2017proximal} which drastically improves \emph{undiscounted training} while achieving similar or better performance than vanilla PPO across our tasks. 
    
    \item We further motivate hierarchical RL (HRL) for improved abstract reasoning, beyond typical sparse reward problems. We show that an HRL approach which exploits a hand-designed abstraction of the problem significantly outperforms standard (flat) deep RL. However, existing general-purpose HRL algorithms, while often claiming to produce meaningful temporal abstractions \cite{frans2017meta, bacon2017option}, fail to provide these benefits. 
\end{itemize}

The rest of the paper is organized as follows. Section 2 provides an overview of related work. In Section 3, we describe our proposed combinatorial RL problems and how to encode them as non-sparse-reward MDPs. In Section 4, we investigate the prospect of solving these combinatorially hard, long-horizon tasks with standard deep RL, focusing on the important role of the discount factor. Section 5 explores whether the temporal abstractions afforded by hierarchy improve reasoning on these tasks. We refer to the background section in the Appendix for preliminary material on RL.


\begin{figure}[t]
    \centering

    \begin{tikzpicture}
    \node at (-4.6,0) {\includegraphics[width=0.32\columnwidth]{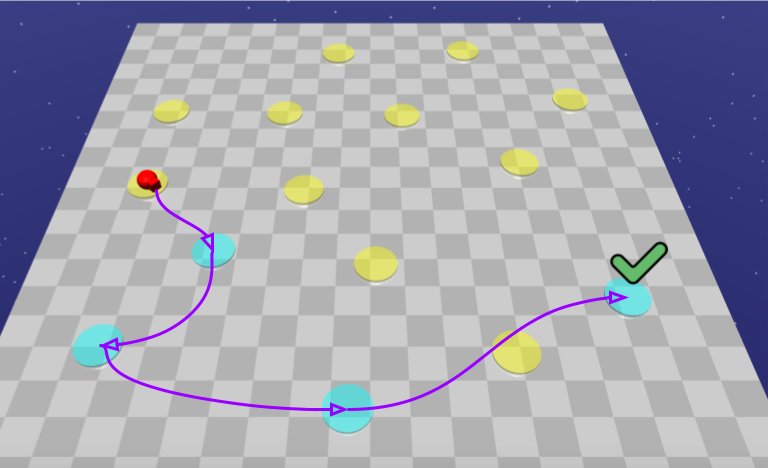}};
    
    \draw (-4.7,-1.8) node {(a) PointTSP};
    
    \node at (0,0) {\includegraphics[width=0.32\columnwidth]{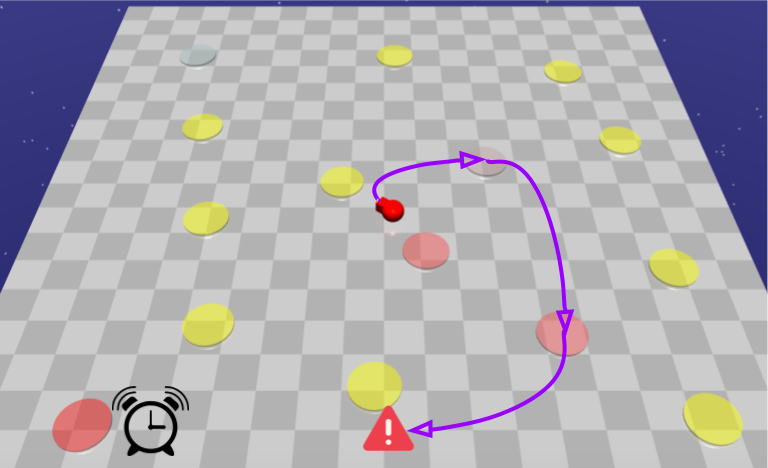}};
    
    \draw (-0.1,-1.8) node {(b) TimedTSP};
    
    \node at (4.6,0) {\includegraphics[width=0.32\columnwidth]{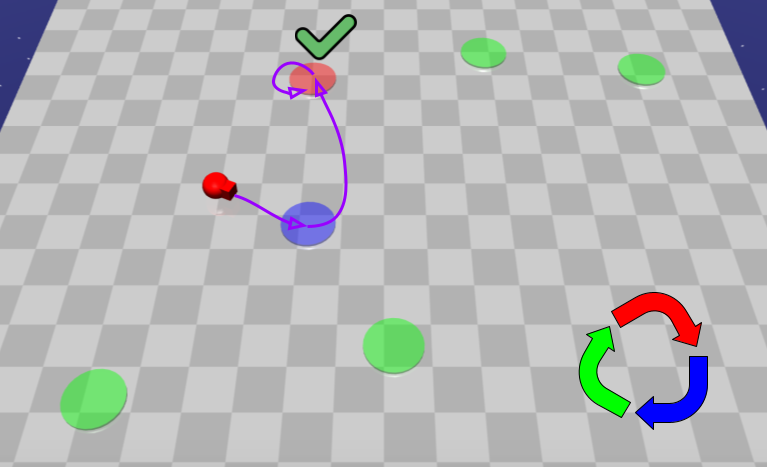}};
    
    \draw (4.5,-1.8) node {(c) ColourMatch};

    \end{tikzpicture}
    \vspace{-7mm}
    
    \caption{Combinatorial problems embedded in a MuJoCo robotics environment. \textbf{(a)} Visit all the zones as quickly as possible (blue = \emph{to be visited}, yellow = \emph{already visited}). A possible path is marked in purple. \textbf{(b)} Visit all zones (or as many as possible), but each zone is initialized with a timeout (darker red = \emph{lower timeout}). Failing to visit any zone before it times out ends the episode (for the purple trajectory, the bottom-left zone timed out). \textbf{(c)} Make all zones the same colour. Visiting a zone cycles its colour in a fixed order (3 possible colours). }
    \label{fig:envs}
\end{figure}
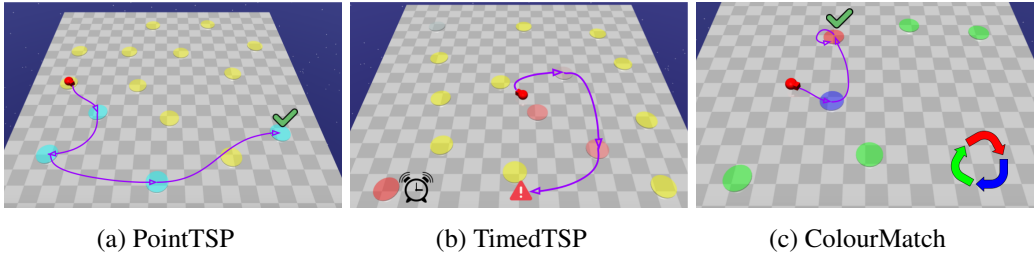



\section{Related Work}

Reinforcement learning has been successfully applied to a wide variety of domains involving complex reasoning. Such domains are often challenging due to a combinatorially large state space. In symbolic domains with known models --- including games (e.g. Chess and Go) \cite{lai2015giraffe,alphazero,anthony2017thinking, pmlr-v97-foerster19a}, theorem proving \cite{kaliszyk2018reinforcement}, symbolic regression \cite{pmlr-v139-landajuela21a}, and combinatorial optimization problems \cite{mazyavkina2021reinforcement} --- methods typically rely on some form of search to improve reasoning. Recent RL research has also targeted discrete, combinatorally hard games with visual observations \cite{bagatella2021planning} such as Sokoban \cite{NIPS2017_9e82757e}.
Unfortunately, combinatorially rich problems in complex, long-horizon environments generally pose significant issues for RL \cite{mirza2020physically} due to sparse rewards, with solutions often requiring additional assumptions, such as domain knowledge \cite{vaezipoor2021ltl2action, leon2020systematic} or expert data \cite{guss2019minerl}. While we are interested more broadly in long-horizon problems that involve complex decision-making, our choice of tasks may evoke similarities to application areas such as robot motion planning \cite{mohanan2018survey, garrett2021integrated}. Indeed, motion planning problems also pose challenges in both long-term planning and low-level control. Traditionally, solutions often relied on domain knowledge 
\cite{latombe2012robot}, but in recent years deep RL solutions have also been proposed \cite{aradi2020survey}.



\newcommand{\rem}{\mathrm{rem}}

\section{Combinatorial RL Tasks}
\label{section:tasks}

In this section, we propose a suite of long-horizon, optimization-based robotics tasks\footnote{Our environments are built off of OpenAI's Safety Gym \cite{Ray2019}.} with underlying combinatorial structure (see Figure \ref{fig:envs}). Solving these tasks well requires the agent to learn both low-level motor control as well as long-term reasoning to optimize the performance criteria. We highlight two key properties of our domains before describing the specific tasks.

\noindent\underline{\textit{Combinatorial solution set:}} Each randomly generated instance from our proposed set of tasks admits a large number of distinct high-level feasible solutions. For example, in the PointTSP task (where the goal is to visit all ``cities" on the grid as quickly as possible), there are $15!$ possible orderings in which the agent can visit the cities. Similar to NP-hard optimization tasks, we expect finding the optimal strategy to be challenging for most instances. 
Note that the optimal strategy depends on a number of factors, e.g., the placement of the cities, the speed and turning radius of the robotic agent, etc. This makes it difficult to handcraft strategies for these tasks.
To ensure that agents cannot simply memorize well-performing solutions to a particular instance, we randomly generate a new map each episode (during training and evaluation), forcing the agent to learn a general strategy.


\noindent\underline{\textit{Reward density:}} In RL task design, it is often the case that rewards become increasingly sparse as we attempt to scale up the task complexity and horizon length  (e.g. \cite{mirza2020physically, duan2016benchmarking, yu2020meta}). While previous works have dealt with reward sparsity through specialized exploration \cite{eysenbach2018diversity, gregor2016variational}, imitation learning \cite{hussein2017imitation}, or curriculum learning \cite{portelas2020automatic}, these methods may negatively bias the learned policy or require additional assumptions.
Instead, dense rewards are a property of our tasks by design, allowing us to train policies via standard RL algorithms, tabula rasa.

\noindent\textbf{PointTSP} Inspired by the NP-hard Travelling Salesman Problem (TSP), the objective of this task is to visit all 15 ``cities'', represented by zones, as quickly as possible (or if this is impossible within the time limit of 2000 steps, to visit as many cities). The robotic agent can accelerate and turn to navigate to these zones and observes its own position, velocity, the positions of zones (and their visitation status) and the time remaining in the episode. We assume for now that rewards are undiscounted (i.e. $\gamma = 1$, an assumption we discuss in Section~\ref{sec:discounting}).
The goal of reaching all 15 zones as quickly as possible is encoded via a sparse reward of $\lambda t_\rem$, where $\lambda \in (0, \infty)$ is a hyperparameter (set to $\lambda=0.01$ in our experiments), and $t_\rem$ is the time remaining when the task is solved. Observe that the reward is Markovian because the time remaining is observable to the agent. A dense reward of $1$ is also provided whenever a new city is reached. 
Note that the dense reward is not misaligned with the original objective, since any successful trajectory garners a total return of $15 + \lambda t_\rem$, while any unsuccessful trajectory garners a less than 15 total return. Thus, the agent is incentivized, first and foremost, to succeed in the task, and secondarily to succeed as fast as possible.

\noindent\textbf{TimedTSP} This task introduces \emph{timeouts} to PointTSP. Each zone $i \in \{1,...,15\}$ is randomly and independently initialized with a timeout $c_i$ from a Beta distribution, observable to the agent. If any zone remains unvisited when its timeout expires, the episode ends and is considered unsuccessful. Thus, the agent may need to visit zones in an inefficient order (from the perspective of PointTSP) 
to prevent a single zone from expiring. Note that in some instances, a zone expiring may be imminent (e.g. if two far-apart zones are both about to expire), in which case it may be optimal to quickly visit as many zones as possible rather than trying to address the expiring zones. 

\noindent\textbf{ColourMatch} This environment contains six zones, each with a randomly initialized (observable) colour: red, green, or blue. Visiting a zone changes its colour to the next colour in a cycle, i.e. green $\rightarrow$ red $\rightarrow$ blue $\rightarrow$ green, and prevents it from being changed again for a short duration. The objective is to make all zones the same colour as quickly as possible. As before, a sparse reward of $\lambda t_\rem$ is provided on the timestep the objective is met, incentivizing the agent to succeed as quickly as possible. To help decompose the task, we define the Hamming distance $H(s)$ in a state $s$ as the minimum number of swaps required to solve the task. We provide a dense reward to the agent each time it changes the colour of a zone: 
for the transition $s \rightarrow s'$ where a zone colour change occurs, the dense reward would be $H(s) - H(s') \in \{1, 0, -1, -2\}$. This dense reward does not bias the original objective, since any successful trajectory from the initial state $s_0$ achieves the same total dense reward of $H(s_0)$. A valid strategy is to only cause colour changes that lead to positive dense rewards, solving the task in the fewest number of \emph{colour swaps}. However, this may not optimally solve the task \emph{as quickly as possible} since it does not consider other relevant factors, e.g., the location of the agent and zones. 

\subsection{Order-invariant Neural Architecture}
\label{subsection:architecture}


Many combinatorial optimization problems are invariant to the order of their inputs, such as the order of jobs in a scheduling problem, or the order of graph edges in a TSP problem. We would also like to exploit this property in our domains. Consider as an example the PointTSP domain: the agent receives observations regarding the position and visitation status of zones, 
and the order in which the observations are conveyed is unimportant. A naive encoding 
that concatenates all state observations and feeds them into (say) a MLP loses this property of order invariance. This leads to sample inefficient learning as an agent does not recognize that a reward accrued by visiting a particular zone could have been gained by visiting \emph{any} zone.

Instead, we encode the order-invariance of zones by borrowing ideas from Deep Set networks. We treat the inputs related to the $K$ zones as an unordered set $\{\mathbf{z}_1, ..., \mathbf{z}_K\}$, where each $\mathbf{z}_i$ is a fixed-sized vector consisting of the position and colour/visitation of that zone. All other aspects of the observation (agent's position and velocity, time remaining, etc) are denoted by a fixed-sized vector $\mathbf{x}$. We use an MLP layer $f$ to jointly encode each zone $\mathbf{z}_i$ with the remaining observation $\mathbf{x}$ which is then aggregated over all zones via averaging, i.e. $\mathbf{w^{(1)}} = \frac{1}{n}\sum_{i=1}^K {f(\mathrm{Concat}(\mathbf{x}, \mathbf{z}_i))}$. We then pass the aggregated vector $\mathbf{w^{(1)}}$ and $\mathbf{x}$ through an additional MLP layer $g$, allowing $g$ to directly attend to relevant information in $\mathbf{x}$ (e.g. the time remaining) which may be difficult to otherwise capture in $\mathbf{w^{(1)}}$, i.e. $\mathbf{w^{(2)}} = g(\mathrm{Concat}(\mathbf{w^{(1)}}, \mathbf{x}))$. $\mathbf{w^{(2)}}$ is then passed directly to the RL agent. All methods in our experiments used this order-invariant neural architecture to encode observations. 

Finally, we remark that other observation modalities may naturally exploit order-invariance, such as images (i.e. the same image observation is generated regardless of the zone ordering). However, learning to process image observations is often a strenuous task in itself and is detrimental to sample efficiency. This is not our main focus in the domains we explore and we instead opted for a lightweight representation of our observations.

\section{The Role of Discounting}
\label{sec:discounting}

With the proposed long-horizon reasoning tasks encoded as MDPs, we now explore the prospect of solving these tasks directly with deep RL. We focus in particular on the discount factor $\gamma \in [0,1]$, which plays a pivotal role by incentivizing immediate rewards over rewards in the future. Discounting is known to improve training stability \cite{amit2020discount, schulman2015high, marbach2003approximate} and is nearly always applied in deep RL, even when the true goal is to maximize an \emph{undiscounted} sum of rewards \cite{andrychowicz2020matters}. Indeed, in our domains, which are finite-horizon (up to 2000 steps), the undiscounted objective is our goal. This assumption is made (in Section~\ref{section:tasks}) to ensure task rewards align with the goal of solving the task as quickly as possible. 

While the benefits of discounting in deep RL are clear, it may unfortunately lead to a mismatch between the desired goal and the optimization objective. For example, when discounting at $\gamma=0.99$, a common discount factor for MuJoCo tasks, a reward of $1$ a thousand steps in the future is reduced to $\approx 0.00004$, while our tasks last up to twice that many steps. To the best of our knowledge, domains demonstrating a significant objective mismatch due to discounting have thus far been limited to simple, tabular examples with known optimal policies \cite{naik2019discounted, mahadevan1996average}.
In more complex domains, the optimal undiscounted return is rarely known due to the instability associated with training an undiscounted objective. 
We hypothesize that this \emph{discounted objective mismatch} is especially critical to address in domains with long-term dependencies since an agent often must forego immediate rewards to maximize its long-term rewards. The following experiments investigate this hypothesis.


\begin{figure}
    \centering
    \begin{tikzpicture}
    \node at (-0.25,0) {\includegraphics[width=0.95\columnwidth]{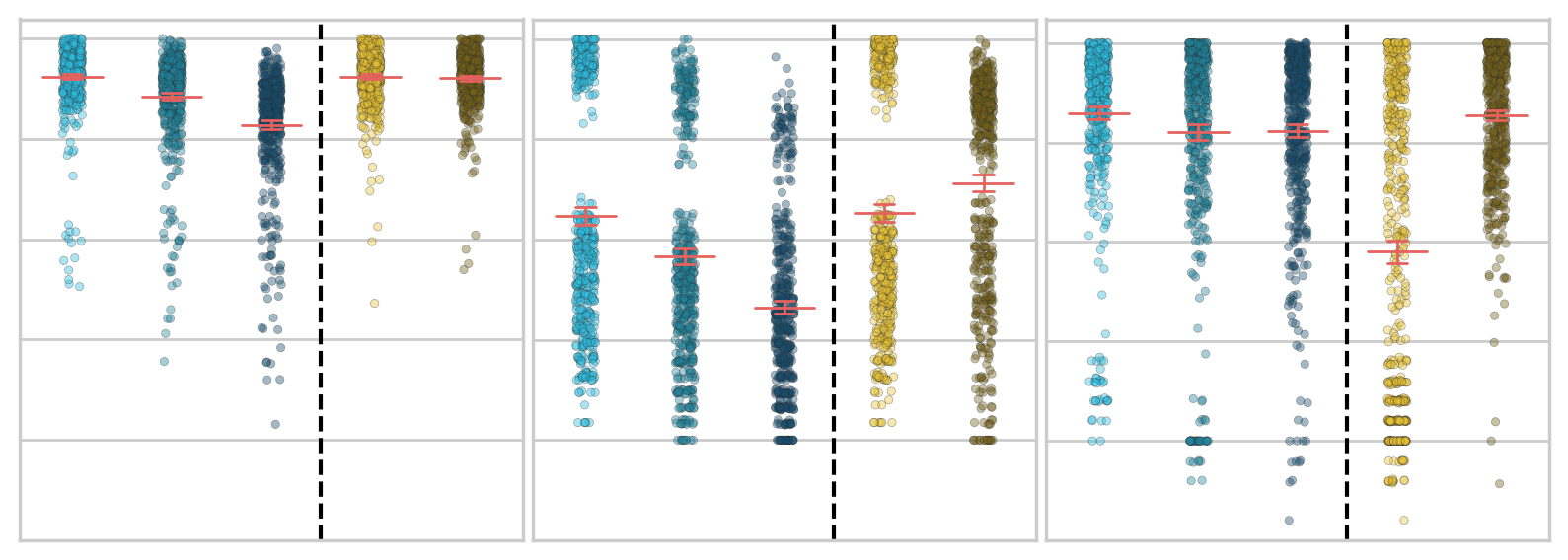}};
    
    \draw (-6.9,0) node[rotate=90] {Normalized Return};
    
    \normalsize
    \draw (-6.75,-2.45) node {$\gamma\!\!=$};
    
    \scriptsize
    \draw (6.45, 2) node {$1.0$};
    \draw (6.45, 1.2) node {$.75$};
    \draw (6.45, 0.4) node {$.50$};
    \draw (6.45, -0.5) node {$.25$};
    \draw (6.45, -1.4) node {$0.0$};
    \draw (6.43, -2.1) node {$-.25$};
    
    \scriptsize
    \draw (-6.3,-2.4) node {$0.99$};
    \draw (-6.3 + 0.9,-2.4) node {$0.9975$};
    \draw (-6.3 + 1.8,-2.4) node {$1$};
    
    \draw (-6.3 + 2.7,-2.4) node {$0.99$};
    \draw (-6.3 + 3.6,-2.4) node {$1$};
    \normalsize
    \draw (-6.3 + 0.9,-2.8) node {PPO};
    \draw [<-] (-6.3 - 0.35,-2.8) -- (-6.3 + 0.5 ,-2.8);
    \draw [->] (-6.3 + 1.3,-2.8) -- (-6.3 + 2.2 ,-2.8);
    
    \draw (-6.3 + 3.2,-2.8) node {\vdn};
    \draw [<-] (-6.3 + 2.2,-2.8) -- (-6.3 + 2.6 ,-2.8);
    \draw [->] (-6.3 + 3.72,-2.8) -- (-6.3 + 3.9 ,-2.8);
    
    \draw (-6.3 + 1.8,2.4) node {PointTSP};

    \scriptsize
    \draw (-2,-2.4) node {$0.99$};
    \draw (-2 + 0.9,-2.4) node {$0.9975$};
    \draw (-2 + 1.8,-2.4) node {$1$};
    
    \draw (-2 + 2.7,-2.4) node {$0.99$};
    \draw (-2 + 3.5,-2.4) node {$1$};
    \normalsize
    \draw (-2 + 0.9,-2.8) node {PPO};
    \draw [<-] (-2 - 0.35,-2.8) -- (-2 + 0.5 ,-2.8);
    \draw [->] (-2 + 1.3,-2.8) -- (-2 + 2.2 ,-2.8);
    
    \draw (-2 + 3.2,-2.8) node {\vdn};
    \draw [<-] (-2 + 2.2,-2.8) -- (-2 + 2.6 ,-2.8);
    \draw [->] (-2 + 3.72,-2.8) -- (-2 + 3.9 ,-2.8);
    
    \draw (-1.9 + 1.8,2.4) node {TimedTSP};
    
    \scriptsize
    \draw (2.4,-2.4) node {$0.99$};
    \draw (2.4 + 0.9,-2.4) node {$0.9975$};
    \draw (2.4 + 1.8,-2.4) node {$1$};
    
    \draw (2.4 + 2.7,-2.4) node {$0.99$};
    \draw (2.4 + 3.4,-2.4) node {$1$};
    \normalsize
    \draw (2.4 + 0.9,-2.8) node {PPO};
    \draw [<-] (2.4 - 0.35,-2.8) -- (2.4 + 0.5 ,-2.8);
    \draw [->] (2.4 + 1.3,-2.8) -- (2.4 + 2.2 ,-2.8);
    
    \draw (2.4 + 3.2,-2.8) node {\vdn};
    \draw [<-] (2.4 + 2.2,-2.8) -- (2.4 + 2.6 ,-2.8);
    \draw [->] (2.4 + 3.72,-2.8) -- (2.4 + 3.9 ,-2.8);
    
    \draw (2.5 + 1.8,2.4) node {ColourMatch};
    \end{tikzpicture}
    \vspace{-6mm}
    
    \caption{Final performances (normalized undiscounted returns) for PPO \& \vdn trained with various discount factors. Returns are averaged over 5 policies $\times$ 100 random instances, then normalized by the best known performance for that instance. Error bars show 90\% confidence intervals.}
    \label{fig:ppo_res}
\end{figure}
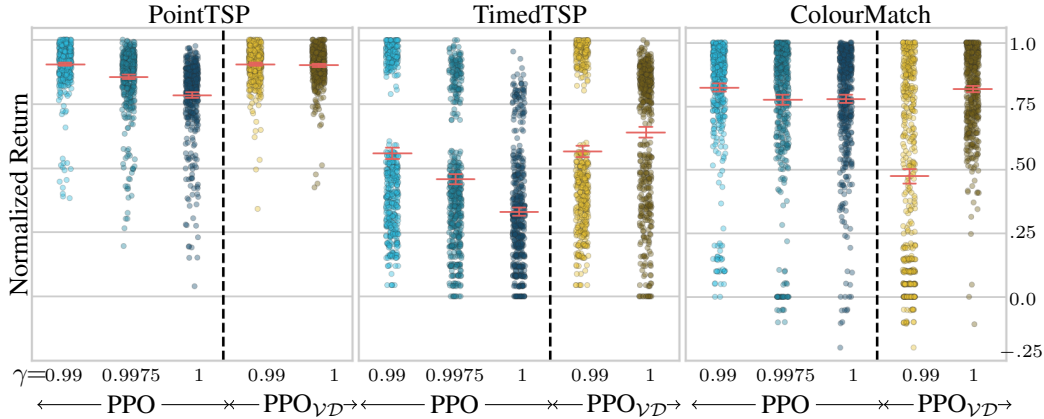

\noindent\textbf{Is discounting necessary for training?} We first verify that discounting indeed improves performance on our proposed tasks, corroborating previous empirical studies that undiscounted training is unstable \cite{andrychowicz2020matters, zhang2020deeper}. We trained PPO with discount factors $\gamma \in \{0.99, 0.9975, 1\}$ on each of the three proposed reasoning tasks. Final performances (as measured by \emph{undiscounted} return) are reported in Figure~\ref{fig:ppo_res}. Training with lower discount factors improved the final performance on all domains, particularly TimedTSP. On PointTSP and TimedTSP, lower discount factors also significantly improved sample efficiency (as evidenced by the learning curves in the Appendix).

\subsection{Training Without Discounting} 
We next discuss why undiscounted RL is so difficult and motivate a simple change to PPO which significantly improves performance in the undiscounted setting.

\noindent\textbf{Value estimation with long horizons.}
We next consider how discounting affects \emph{value estimation} with function approximation. In policy gradient methods, estimated values play a critical role in bootstrapping $n$-step returns and reducing variance as a baseline. Formally, we consider estimating the value function $V^{\pi, \gamma}(s) = \mathbb{E}_{\tau \sim \pi | s_0=s}[ G^\gamma (\tau) ]$, where $G^\gamma(\tau) = \sum_{i=1}^{H} {\gamma^{i-1}r_i}$ is the return of a trajectory $\tau$ (with rewards $r_1,...,r_H$) under discount factor $\gamma$. 


Due to stochasticity in the policy $\pi$ and/or the environment dynamics, the return $G^\gamma(\tau)$ of a random trajectory starting from state $s$ can have high variance. We emphasize that this variance often depends on the remaining horizon of $s$ --- states $s$ with many remaining steps in the episode tend to have higher uncertainty in their returns. This can potentially be problematic at training time for several related reasons: (1) accurately estimating $V^{\pi, \gamma}(s)$ requires more samples; (2) the learned value function is more prone to overfitting to a poor batch of data, leading to instability; (3) states with high variance returns (e.g., those at the beginning of a long episode) may dominate the value loss, worsening the quality of predicted values for other states. 

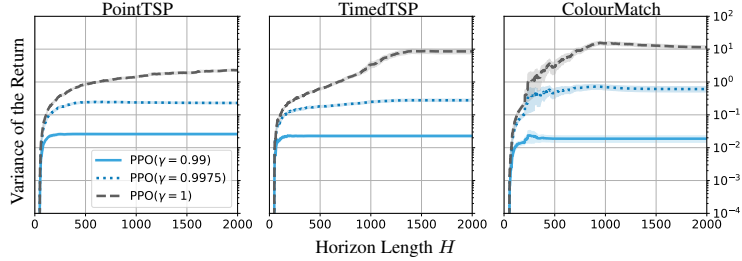
\begin{figure}
    \centering
    \begin{tikzpicture}
    \node at (0,0) {\includegraphics[width=0.7\columnwidth]{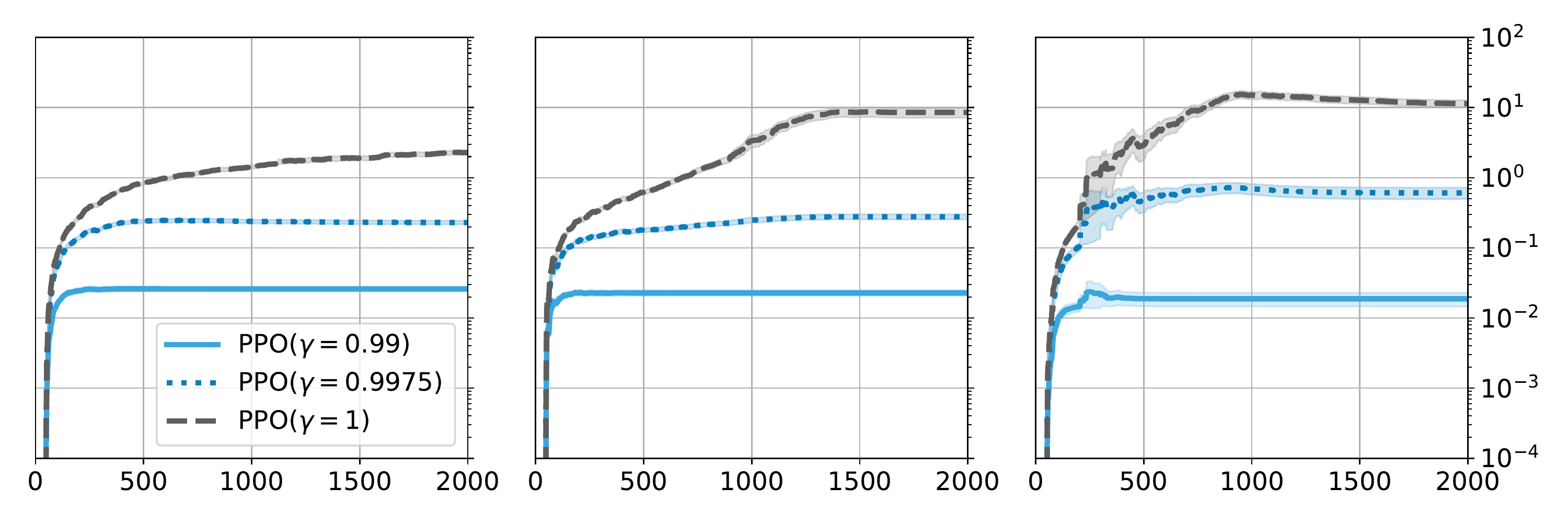}};
    \scriptsize
    \draw (-3.3,1.5) node {PointTSP}; 
    \draw (-0.1,1.5) node {TimedTSP}; 
    \draw (3,1.5) node {ColourMatch}; 
    
    \draw (-4.9,0) node[rotate=90] {Variance of the Return};
    \draw (0,-1.7) node {Horizon Length $H$};
    
    \end{tikzpicture}
    \vspace{-3mm}
    
    \caption{Variance of the return drastically increases with the horizon length in the undiscounted case. We show empirical variance in the return $G^\gamma(\tau)$ from 20 sampled trajectories of a fixed policy and initial state, for various horizon lengths and $\gamma$. Results are averaged over 20 instances~$\times$~5 policies, with error bars showing standard error. Note the log scale on the $y$-axis.}
    \label{fig:env_variances}

\end{figure}

We demonstrate the severity of this variance in our environments. Figure~\ref{fig:env_variances} shows the empirical variance of $G^\gamma(\tau)$ from rolling out a fixed policy (obtained partway through training) from the same initial state 20 times. When evaluating $G^\gamma(\tau)$, we vary the discount factor $\gamma \in \{0.99, 0.9975, 1\}$ as well as the horizon length of the episode $H \in \{1,2,...,2000\}$. With $\gamma=1$, the variance of the return $G^\gamma(\tau)$ drastically increased with the horizon length while discounting with $\gamma=0.99$ maintained a relatively low variance throughout the episode. This is expected as discounting effectively shortens the horizon by reducing the contributions of distant future rewards.


\noindent\textbf{Dealing with high variance in value estimation.} 
Besides discounting, value bootstrapping methods such as GAE \cite{schulman2015high} are commonly used to trade-off between bias and variance in long-horizon problems. Unfortunately, even with such techniques, PPO trained poorly in the undiscounted setting. 

Based on our previous observations, we propose and evaluate a simple modification to PPO to ameliorate the issue of high variance in $G^\gamma(\tau)$ when training without discounting.
We consider learning the mean and variance of $G^\gamma (\tau)$, rather than a point estimate of the mean (as is typically done by the value network in PPO) for any state. This approach, which we refer to as \vdn ($\mathcal{VD}$ for "value distribution") is described in Algorithm~\ref{algo:ppovd}. Intuitively, for states $s$ where the return is predicted to have high variance, \vdn performs a more conservative update compared to a point estimate of the value, making \vdn less prone to overfitting within a batch of experience. While \vdn has similar motivations to distributional RL \cite{bellemare2017distributional, nam2021gmac}, 
we opt for a simpler setup, requiring minimal changes to the standard PPO algorithm. For example, we model the value distribution as Gaussian, which only requires learning a scalar mean and standard deviation, and the $n$-step returns (which use predicted values for bootstrapping) are treated as scalar targets rather than distributions. We are unaware of any prior work which learns value distributions to specifically address the instability of high discount factors in long-horizon environments. 

\begin{algorithm}[t]
    \SetAlgoLined
    
    \textbf{Inputs:} Number of steps $T$, discount factor $\gamma$, GAE parameter $\lambda$, number of optimization epochs $K$, policy network parameters $\theta$, value network parameters $\psi$ \;
    
    \vspace{3mm}
    \tcc{Data collection + processing}
    Collect $T$ on-policy samples $(s_0, a_0, r_1, s_1, ..., a_T, r_T, s_{T+1})$ using $\pi_\theta$\;
    Compute estimated values $v_t = \mathrm{value\_net}_\psi(s_t).\mathrm{{\color{red} mean}}$\; 
    Compute estimated advantages $\mathrm{adv}_t = \sum_{l=0}^{T-t}{(\gamma\lambda)^l (r_{t+l} + \gamma v_{t+l+1} - v_{t+l})}$\;
    Compute value targets $\hat{v}_t = v_t + \mathrm{adv}_t$\;
    \vspace{3mm}
    \tcc{Policy optimization}
    \For{$K$ epochs}{
        Update $\theta$ using PPO loss and advantages\; 
        Update $\psi$ using {\color{red} Gaussian negative-log-likelihood loss and value targets}\; 
    }
    \caption{\vdn (1 iteration). Changes with respect to standard PPO are marked in \color{red}{red}.}
    \label{algo:ppovd}
\end{algorithm}

The performance of \vdn is reported in Figure~\ref{fig:ppo_res}. A version of \vdn with $\gamma=0.99$ is also included for comparison, though the algorithm is motivated for the undiscounted case.
\vd{1} converged to similar or better performance than PPO with discounting across all domains. Notably, on TimedTSP, \vd{1} significantly outperformed \ppo{0.99} which we believe is due to optimizing the true objective -- we provide further evidence of this below. Furthermore, \vd{1} significantly outperformed \ppo{1}, which we attribute to the improved stability of our approach in the undiscounted case. As expected, however, instability does not appear as problematic in the discounted setting: modelling a value distribution (i.e. \vd{0.99}) provided no additional benefit on top of \ppo{0.99} and was sometimes detrimental to performance. 
Lastly, discounting appears advantageous for sample efficiency: \ppo{0.99} was generally more sample efficient than \vd{1}.



\begin{figure}[tb]
    \centering
    \begin{tikzpicture}
    \node at (-0.12\columnwidth,0) {\includegraphics[width=0.27\columnwidth]{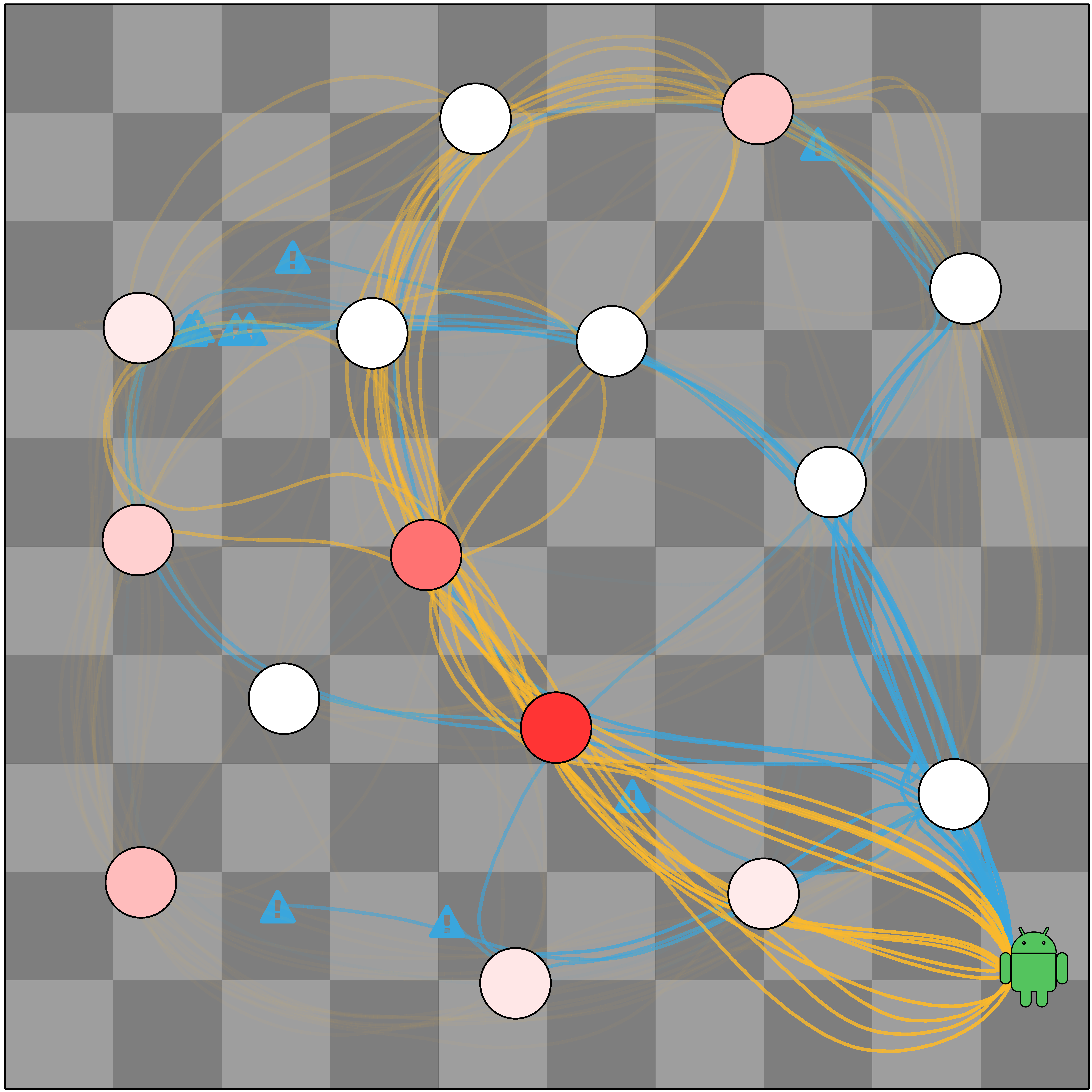}};
    \node at (0.28\columnwidth,-0.23) {\includegraphics[height=0.31\columnwidth]{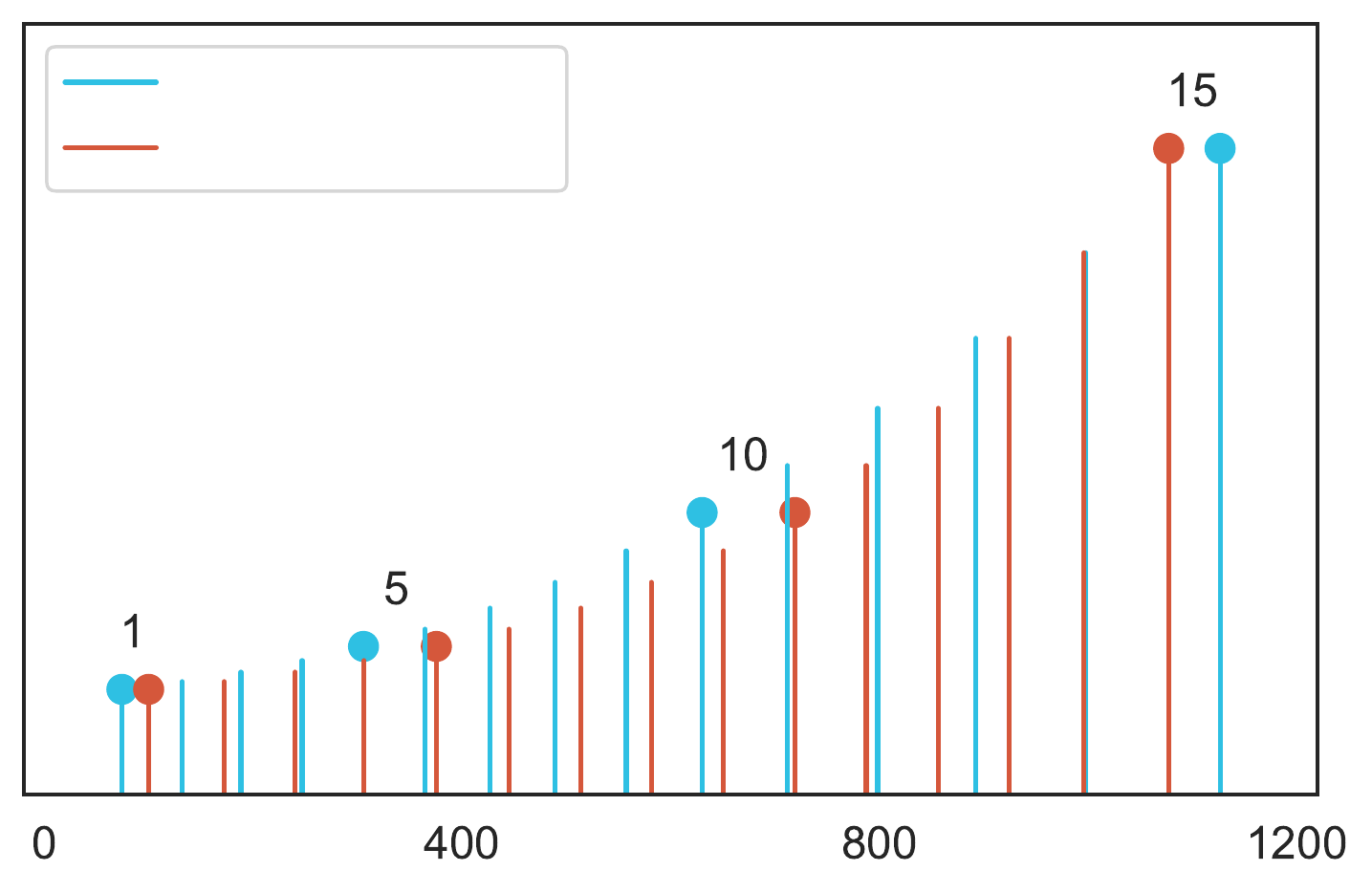}};
    \small
    \draw (2.3,1.55) node {\ppo{0.99}}; 
    \draw (2.2,1.15) node {Zone-goals}; 
    
    \draw (0.29\columnwidth,-2.5) node {Number of timesteps};
    \end{tikzpicture}
    \vspace{-3mm}

    \caption{
    \textbf{(left:)} A visualization of 15 trajectories ($5$ policies $\times$ $3$ rollouts) from \ppo{0.99} (\emph{blue}) and \vd{1} (\emph{orange}) on a TimedTSP instance. Circles indicate zones, with darker red indicating lower timeout; the green robot indicates the starting position; X's indicate the end of a trajectory due to a timeout. Trajectories become more transparent over time to focus on each policy's initial behaviour. \vd{1} immediately visits 
    the two zones with lowest timeout, while \ppo{0.99} neglects this and quickly fails. \textbf{(right:)} The cumulative time spent in PointTSP to visit $i$ zones, for $i \in \{1,2,...,15\}$ (averaged over $5$ policies $\times$ 100 instances). The height of each line increases with $i$, and lines corresponding to $i\in \{1,5,10,15\}$ are marked with a circle. \ppo{0.99} rapidly visits the first few zones, but spends more time visiting all 15 zones than Zone-goals.}
    \label{fig:sketches}
    \vspace{-0.45cm}
\end{figure}
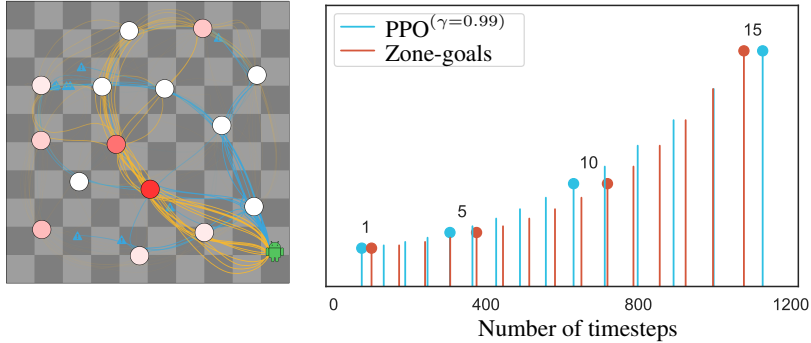

\noindent\textbf{How does objective mismatch due to discounting affect performance?} 
We previously observed that optimizing the \emph{undiscounted} objective with \vdn can improve performance over discounted RL. Here we take a closer look at how discounting can be detrimental to decision-making. 

We report in Table~\ref{objective-mi-table} the final performances of \ppo{0.99}, \vd{1}, and Zone-goals$^{(\gamma=1)}$ (a strong baseline we introduce in the next section to optimize the undiscounted objective) using both discounted ($\gamma=0.99$) and undiscounted ($\gamma=1$) returns as the evaluation criteria. We find evidence of objective mismatch due to discounting on TimedTSP and, to a lesser extent, on PointTSP --- Zone-goals simultaneously performed better than \ppo{0.99} when evaluated with $\gamma=1$ but also worse when evaluated with $\gamma=0.99$.

\begin{wraptable}{r}{8cm}
\caption{On PointTSP and TimedTSP, the best baseline depends on $\gamma_{\text{eval}}$ showing that the $\gamma=1$ and $\gamma=0.99$ objectives are not aligned. The average return of 500 ($5$ policies $\times$ $100$ instances) trajectories from each approach is reported.}
\label{objective-mi-table}
\centering
\scriptsize
\begin{tabular}{lrcccccc}

    \toprule
                  &   & \multicolumn{2}{c}{PointTSP} & \multicolumn{2}{c}{TimedTSP} & \multicolumn{2}{c}{ColourMatch} \\ 
    \cline{3-8}
    baseline & $\gamma_{\text{eval}}$ \!\!\! & 0.99          & 1             & 0.99          & 1              & 0.99           & 1              \\ \hline
    \multicolumn{2}{l}{\ppo{0.99}}              & \textbf{1.13} & 23.48 & \textbf{1.12} & 14.15   & 0.60    & 16.45   \\
    \multicolumn{2}{l}{\vd{1}}           & 0.66   & 23.36  & 0.67    & 16.19   & 0.32    & 16.33   \\
    \multicolumn{2}{l}{Zone-goals$^{(\gamma=1)}$}       & 0.81  & \textbf{24.24}  & 0.91  & \textbf{21.95}   & \textbf{0.93}    & \textbf{18.95}  \\
    \bottomrule
    \vspace{-5mm}
\end{tabular}
\end{wraptable}

%




On TimedTSP, we observed critical differences in the decision-making of \ppo{0.99} vs \vd{1}. A visualization of each method's behaviour on a challenging instance is shown in Figure~\ref{fig:sketches} (left). In this instance, two zones are initially close to timing out and must be visited promptly to continue the episode. All trajectories from \vd{1} immediately visited these zones and managed to achieve a high return while most trajectories from \ppo{0.99} neglected these zones and quickly failed. Notably, \ppo{0.99} always greedily visited the closest zone to the agent's starting position first, while none of the \vd{1} trajectories did. 


While less pernicious towards performance, discounted PPO also showed signs of myopic decision-making on PointTSP. We observed that \ppo{0.99} often left outlying cities unvisited until the end in order to more rapidly visit cities at the start. This is evidenced in Figure~\ref{fig:sketches} (right), where we report the average cumulative time taken to visit $i$ cities (for each $i$ up to 15) for \ppo{0.99} and Zone-goals. Despite visiting the first few cities quicker than Zone-goals, \ppo{0.99} spends considerable time visiting the last few cities and ultimately is slower in solving the task. 

On ColourMatch, we remark that Zone-goals outperformed \ppo{0.99} on both the discounted \emph{and} undiscounted objectives, implying that discounting cannot be the only source of suboptimality. We believe this is since ColourMatch presents a more difficult reasoning task -- in fact, even determining whether visiting a zone will result in a positive reward is non-trivial.



\section{The Role of Hierarchy}

In the previous section, we observed significant issues with PPO in long-horizon reasoning tasks: training with $\gamma$ close to 1 was often unstable while lowering the discount factor exhibited myopic behaviour. Here, we turn our attention to hierarchical reinforcement learning (HRL), which has widely been applied to long-horizon tasks in the past. HRL is often motivated by the hierarchical structure present in real-world tasks --- to make pizza, one must roll out the dough, prepare the sauce, add toppings, and bake the pizza. Unfortunately, as tasks increase in length and complexity, they quickly become unsolvable by methods which learn from random initial behaviour in the sparse reward setting. Many works in deep HRL aim to address this reward sparsity by improving exploration (e.g. \cite{eysenbach2018diversity, li2019hierarchical, florensa2017stochastic, zhang2021hierarchical}). Indeed, mitigating the impact of sparse rewards has been cited as the main benefit resulting from hierarchy in existing methods \cite{nachum2019does, jong2008utility}. 

Here, we attempt to motivate a different rationale for the use of hierarchy. In our long-horizon environments, single actions are unlikely to lead to meaningfully different states and therefore, exploiting the high-level structure in our tasks may require reasoning over concepts that are temporally abstracted.
In this section, we investigate whether the temporal abstractions afforded by HRL indeed improve reasoning on combinatorial RL tasks compared to non-hierarchical RL.






\subsection{Experimental Setup}

A wide variety of HRL algorithms have previously been proposed in the literature and we attempt to cover a representative sample of them here. We restrict ourselves to the two-level hierarchical framework, as is most common in deep RL environments. In this framework, a high-level policy operates at a lower temporal frequency by selecting high-level actions which persist for many timesteps (either a fixed number, or based on state-dependent termination criteria). A low-level policy conditions on the current high-level action to select actions in the environment every timestep. 
We assume both high and low-level policies receive the full environment observation.
The manner in which the high and low-level policies are trained also varies between algorithms. In each of our HRL implementations, we concurrently train both the high and low-level policies with PPO, using the same on-policy data to optimize each, similar to \cite{li2019hierarchical}. While off-policy HRL approaches exist \cite{nachum2018data} and are potentially more sample efficient, they are generally less stable due to the worsened effects of non-stationarity from training both policies simultaneously. All methods were trained for 100 million frames on PointTSP and 150 million frames on TimedTSP and ColourMatch, and the hidden layer sizes were adjusted such that each method used roughly the same number of parameters. The complete hyperparameter settings are reported in the Appendix.

\subsection{Methods and Results}
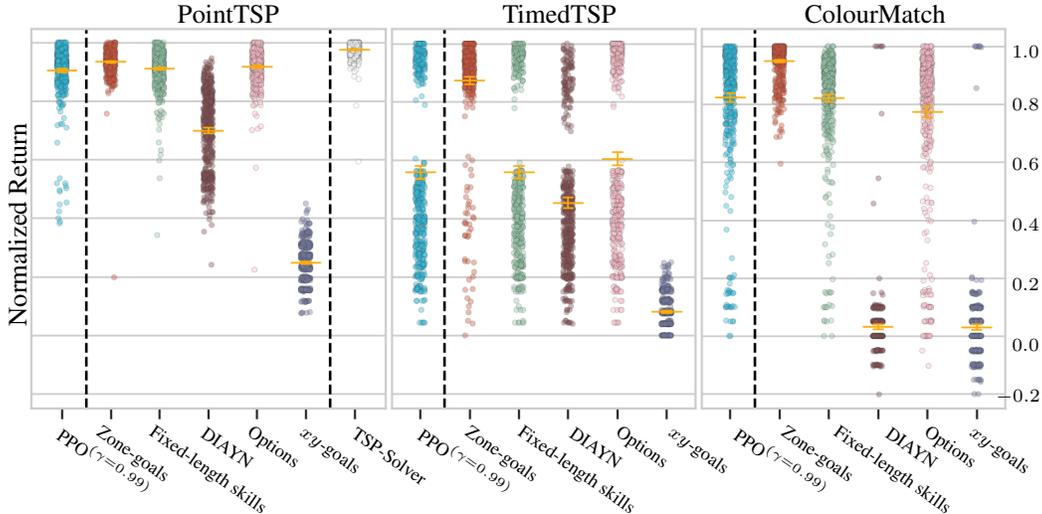
\begin{figure}
    \centering
    \begin{tikzpicture}
    \node at (-0.15,0) {\includegraphics[width=0.95\columnwidth]{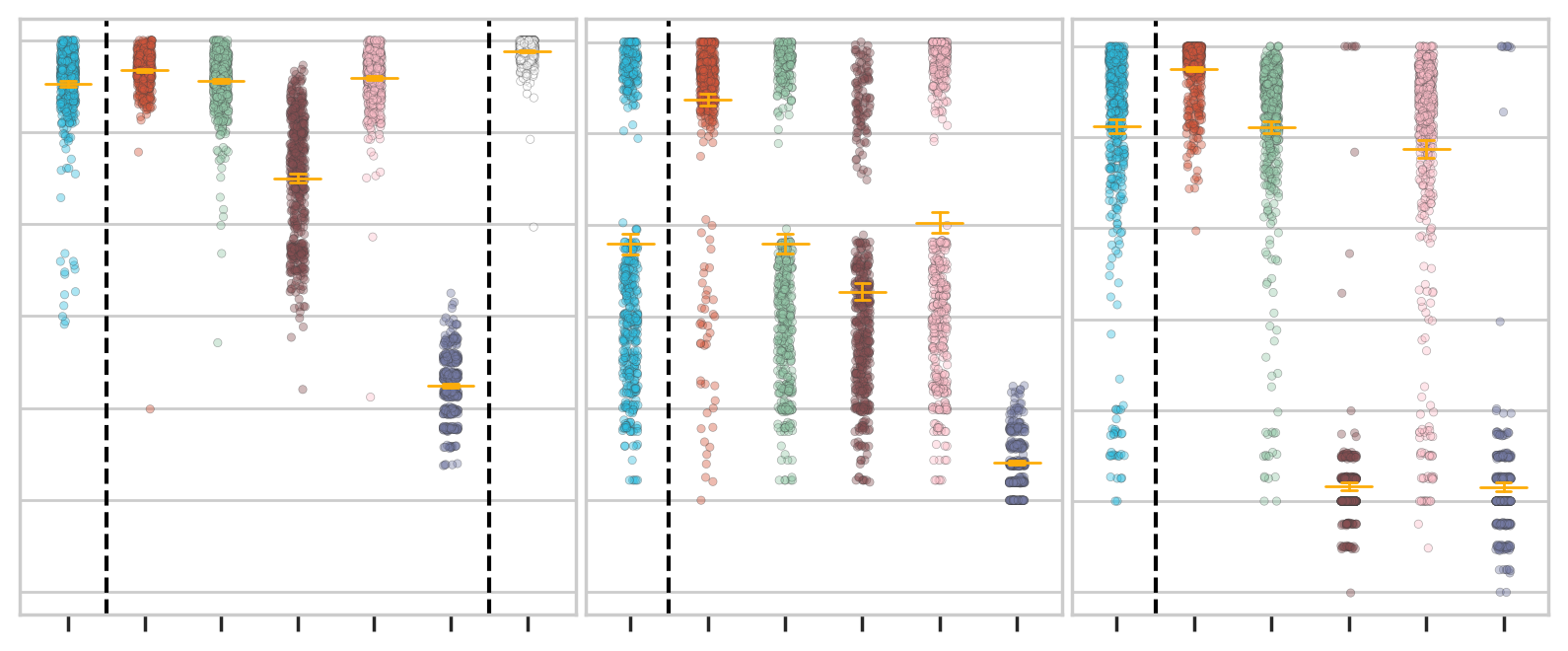}};
    
    \draw (-6.8,0) node[rotate=90] {Normalized Return};
    
    \scriptsize
    \draw (6.6, 2.3) node {$1.0$};
    \draw (6.6, 1.6) node {$0.8$};
    \draw (6.6, 0.8) node {$0.6$};
    \draw (6.6, 0) node {$0.4$};
    \draw (6.6, -0.8) node {$0.2$};
    \draw (6.6, -1.6) node {$0.0$};
    \draw (6.5, -2.3) node {$-0.2$};
    
    \scriptsize
    \draw (-6.4 + 0.1,-2.7) node[rotate=-35, anchor=west] {\ppo{0.99}};
    \draw (-6.4 + 0.6,-2.7) node[rotate=-35, anchor=west]
    {Zone-goals};
    \draw (-6.4 + 1.3,-2.7) node[rotate=-35, anchor=west] {Fixed-length skills};
    \draw (-6.4 + 2,-2.7) node[rotate=-35, anchor=west] {DIAYN};
    \draw (-6.4 + 2.6,-2.7) node[rotate=-35, anchor=west] {Options};
    \draw (-6.4 + 3.3,-2.7) node[rotate=-35, anchor=west] {$xy$-goals};
    \draw (-6.4 + 4,-2.7) node[rotate=-35, anchor=west] {TSP-Solver};
    \normalsize
    \draw (-6 + 2,2.8) node {PointTSP};
    
    \scriptsize
    \draw (-1.5 ,-2.7) node[rotate=-35, anchor=west] {\ppo{0.99}};
    \draw (-1.5 + 0.6,-2.7) node[rotate=-35, anchor=west] {Zone-goals};
    \draw (-1.5 + 1.3,-2.7) node[rotate=-35, anchor=west] {Fixed-length skills};
    \draw (-1.5 + 2,-2.7) node[rotate=-35, anchor=west] {DIAYN};
    \draw (-1.5 + 2.6,-2.7) node[rotate=-35, anchor=west] {Options};
    \draw (-1.5 + 3.3,-2.7) node[rotate=-35, anchor=west] 
    {$xy$-goals};
    \normalsize
    \draw (-1.6 + 2,2.8) node {TimedTSP};
    
    \scriptsize
    \draw (2.7 ,-2.7) node[rotate=-35, anchor=west] {\ppo{0.99}};
    \draw (2.7 + 0.6,-2.7) node[rotate=-35, anchor=west] {Zone-goals};
    \draw (2.7 + 1.3,-2.7) node[rotate=-35, anchor=west] {Fixed-length skills};
    \draw (2.7 + 2,-2.7) node[rotate=-35, anchor=west] {DIAYN};
    \draw (2.7 + 2.5,-2.7) node[rotate=-35, anchor=west] {Options};
    \draw (2.7 + 3.1,-2.7) node[rotate=-35, anchor=west] {$xy$-goals};
    \normalsize
    \draw (2.6 + 2,2.8) node {ColourMatch};
    \end{tikzpicture}
    \vspace{-5mm}
    
    \caption{Final performances (normalized undiscounted returns) for various HRL methods. Returns are averaged over $5$ policies $\times$ $100$ random instances, then normalized by the best known performance for that instance. Error bars show $90\%$ confidence intervals.}
    \label{fig:hrl_res}
    \vspace{-5mm}
    
\end{figure}

We describe the HRL methods we considered and report their performances on our suite of long-horizon reasoning tasks in Figure \ref{fig:hrl_res}.

\noindent\textbf{Fixed-length skills \cite{florensa2017stochastic, frans2017meta}:} A simple and common HRL approach uses the high-level policy to choose a discrete low-level skill to execute every $k$ timesteps, where $k$ is a constant. 
Both the high and low-level policies are optimized directly using the environment reward. We choose to optimize the low-level policy using \emph{discounted} environment rewards as learning was unstable with undiscounted rewards. However, as the high-level policy essentially faces a shorter horizon length due to actions being temporally extended, we optimize the high-level policy using \emph{undiscounted} environment rewards to avoid the issue of objective mismatch.

\textbf{\emph{Result.}} Fixed-length skills achieved similar performance to flat PPO. Even when evaluated with random skills, the performance did not drop, suggesting that the different skills became indistinguishable from one another. In such cases, the HRL policy effectively collapses into a flat policy and is unable to benefit from temporally extended actions.

\noindent\textbf{DIAYN \cite{eysenbach2018diversity}} To avoid learning a homogeneous set of skills, we consider applying an information-theoretic diversity objective to the fixed-length skills framework. 
Such approaches typically learn task-agnostic skills in a reward-free setting to improve exploration. 
Instead, we explore whether promoting skill diversity in our domains can aid in learning more meaningful high-level actions. We linearly combine task rewards with the DIAYN objective when training the low-level policy: $r_t' = r_t + \alpha (\log{q_\phi(z_t | s_{t+1}) - \log{p_\psi(z_t)}})$, where $\alpha \in [0,\infty)$ is a hyperparameter, $r_t$ the environment reward, $z_t$ the current discrete skill, $q_\phi$ a neural network trained to predict the skill distribution given the next state, and $p_\psi$ a neural network trained to predict the prior skill distribution. The high-level policy was trained to maximize undiscounted environment rewards (without the diversity objective).


\textbf{\emph{Result.}} The inclusion of the diversity term performed worse than fixed-length skills in all tasks. Thus, simultaneously optimizing for diversity and reward appears detrimental to reward.

\noindent\textbf{Options \cite{sutton1999between}:} An \emph{option} $\omega$ is a temporally extended action which consists of a low-level policy $\pi_\omega: \mathcal{S} \times \mathcal{A} \rightarrow [0,1]$, a set of states $\mathcal{I}_\omega \subseteq \mathcal{S}$ where $\omega$ can be initiated, and a termination condition $\beta_\omega: \mathcal{S}\rightarrow [0,1]$.  When an option terminates, a high-level policy chooses the next option from among those that can be initiated. To simplify learning, we assume all options can be initiated in any state (similar to \cite{bacon2017option}). We learn the termination condition by augmenting the action-space of the low-level policy with a special $\mathtt{stop}$ action. As before, the low-level policy and the high-level policy over options are trained on the discounted and undiscounted (resp.) environment rewards. 

\textbf{\emph{Result.}} Learning the termination condition 
provided similar performance 
to fixed-length skills.

\noindent\textbf{\emph{xy}-goals:} Inspired by goal-based HRL \cite{gurtler2021hierarchical, nachum2018data, chane2021goal, vezhnevets2017feudal}, we evaluate an approach where the high-level policy assigns $xy$-goals $g \in \mathbb{R}^2$ to the low-level agent. The low-level policy receives only a dense reward based on the change in Euclidean distance to the goal every timestep while the high-level policy directly receives rewards from the environment. 

\textbf{\emph{Result.}} This approach performed poorly on all domains. While the agent was proficient at navigating to the target location, it struggled to even solve simple parts of the task, such as reaching new zones in PointTSP. We note that neither increasing the frequency of goal selection nor using full states as goals (similar in spirit to \cite{nachum2018data}) improved performance. 

\noindent\textbf{Zone-goals (Ours):} We design a domain-specific goal space based on the zones in the environment. With the current observation, the high-level policy produces a score for each zone and selects the next zone the agent should visit using a softmax over the scores. Certain zones in PointTSP and TimedTSP are invalid choices if previously visited, and are masked out, with the policy gradients adjusted accordingly \cite{tang2020implementing}. The low-level policy receives the $xy$-coordinates of the goal zone in addition to the current observation, and is rewarded in the same manner as the $xy$-goals approach. 

\textbf{\emph{Result.}} This method significantly outperformed the previous approaches (including standard PPO), particularly on TimedTSP and ColourMatch. This demonstrates the efficacy of hierarchy towards long-term reasoning, given the proper abstraction of the problem. Despite optimizing the undiscounted reward at the high-level, the instability observed in standard PPO in Section~\ref{sec:discounting} did not present here. 

\noindent\textbf{TSP-Solver (Ours):} On the PointTSP domain, we consider generating a open-loop, high-level plan at the start of every episode over the order in which zones should be visited. The plan is generated using a metric TSP solver with the $xy$-coordinates of zones as city locations. With this fixed ordering of zones (which replaces the role of the high-level policy), the low-level policy is rewarded based on $xy$-distance towards the next unvisited zone. However, we allow the low-level policy to plan ahead by observing the entire ordering of zones --- for the $i$-th zone in the ordering, the observation for that zone is augmented with a scalar value of $2^{-i+1}$ to uniquely identify its position in the ordering. 

\textbf{\emph{Result.}} Using a planner outperformed all other approaches on PointTSP, showing the advantage of abstract planning in complex, long-horizon domains. We note that the plan generated is not necessarily optimal, since the solver does not consider the agent's acceleration, turning speed, and other relevant factors to the problem.

\section{Conclusion}
In this work we study long-horizon robotics tasks that are combinatorially hard but that afford a number of feasible solutions. 
To this end, we propose a representative set of tasks that we attempt to solve using state-of-the-art RL methods.
Our tasks involve both low-level motor control as well as complex reasoning about delayed rewards and distinct possibilities over future states. However, crucially these tasks can be solved (albeit suboptimally) by standard RL methods without specialized exploration.
Through our investigation, we uncovered several weaknesses with state-of-the-art RL algorithms in the context of long-term reasoning, highlighting promising directions for future work.

\noindent\textbf{$\bullet$ Undiscounted RL} Discounting is nearly always applied in long-horizon deep RL problems for stability, however this can cause a significant objective mismatch in some of our domains. We showed that this leads to ``myopic'' decisions which fail to consider delayed rewards. Thus far,
undiscounted RL (or average-reward RL) has received little attention from the deep RL community. We believe that improving the robustness and sample efficiency for undiscounted RL will become increasing important as RL scales to longer and more complex tasks.

\noindent\textbf{$\bullet$ Hierarchical RL} While the benefits of hierarchy in planning and reasoning are often espoused \cite{erol1994umcp, sutton1999between, CURRIE199149}, most HRL research to date has focused on the challenge of exploration. Instead, we motivate the use of hierarchy as a way to improve long-term reasoning and stability where standard RL methods fail. We demonstrated in our combinatorial RL domains that leveraging the proper hierarchical abstraction can substantially improve convergence performance over standard RL. However, this hierarchical abstraction was built using domain knowledge of the problem, and existing general-purpose HRL approaches were unable to provide similar benefits over PPO. This presents opportunities for future research to improve hierarchical reasoning, planning, and discovery beyond sparse reward problems. 

In conclusion, we presented long-horizon RL problems that involve combinatorial reasoning to scale up task complexity while avoiding the issue of sparse rewards, which has traditionally plagued RL task design. We believe these types of problems can expedite future RL research due to the unique challenges they pose to existing RL methods. 

\bibliography{main.bib}
\bibliographystyle{plain}





\newpage
\setcounter{page}{1}
\appendix
\appendixpage
\startcontents[sections]
\printcontents[sections]{l}{0}{\setcounter{tocdepth}{2}}




\section{Background}

In this paper, we consider the discounted formulation of a Markov Decision Process (MDP) $\mathcal{M} = \langle S,T,A,p,r,\gamma,\mu \rangle$, where $S$ is a set of \emph{states}, $T \subseteq S$ a set of \emph{terminal states}, $A$ a set of \emph{actions}, $p(s'|s,a)$ the \emph{transition probability distribution}, $r(s,a,s')$ the \emph{reward function}, $\gamma$ the \emph{discount factor}, and $\mu$ the \emph{initial state distribution}. For each \emph{episode}, the initial state $s_0$ is randomly sampled from $\mu$. The agent then chooses actions $a_t$ at each timestep $t \ge 0$ according to its policy $\pi(a_t | s_t)$ and receives the next state $s_{t+1}$ according to transition probabilities (also known as the \emph{dynamics}) $p(s_{t+1} | s_t, a_t)$ and a reward $r_{t+1}$ according to $r(s_t,a_t,s_{t+1})$. This proceeds until a terminal state in $T$ is reached. The \emph{discounted return} in an episode of length $H$ is defined as $\sum_{i=1}^{H} \gamma^{i-1} r_i$ and the goal of the agent is to learn a policy $\pi(a|s)$ that maximizes the expected discounted return.

We focus in particular on \emph{on-policy policy gradient methods}, a popular class of RL algorithms for environments with continuous actions and states. Here, the policy $\pi_\theta(a|s)$ is stochastic and parametrized by $\theta$ (e.g. $\theta$ could be the weights of a neural network). On-policy methods sample data from the environment according to the current policy $\pi_\theta(a|s)$ and attempt to directly estimate the gradient of the expected return $\nabla_\theta \mathbb{E}[ \sum_{i=1}^{H} \gamma^{i-1} r_i ]$, where the expectation is over the stochasticity of the policy and the environment transitions. The estimated gradients are then used to update the policy parameters $\theta$ (hopefully resulting in a higher expected return) using gradient ascent.

\section{Packages and Licensing}
The code for our proposed set of environments, along with scripts for training and evaluation, are available at \url{https://github.com/andrewli77/combinatorial-rl-tasks}. Our environments are built off of the OpenAI Safety Gym\footnote{\url{https://github.com/openai/safety-gym}} while the RL training code is built off of the torch-ac repository\footnote{\url{https://github.com/lcswillems/torch-ac}}, both of which are available under an MIT license. Note that the OpenAI Safety Gym uses MuJoCo \footnote{\url{https://www.roboti.us/index.html}} which is freely available under an Apache 2.0 license. 

\section{Additional Experimental Details}
All experiments were conducted on a compute cluster using a single GPU and 16 CPU cores per run. PPO training lasted roughly 18 hours for PointTSP (for 100 million frames), 55 hours for PointTSP (for 150 million frames), and 27 hours for ColourMatch (for 150 million frames). Other methods required a similar amount of training time, except for Options, which required about 50\% more time. Note that with fixed-length skills, calls to the high-level policy could be synchronized across parallel training environments, while this was not possible with variable-length Options. 

\begin{figure}
    \centering
    \begin{tikzpicture}
    \node at (0,0) {\includegraphics[width=0.975\columnwidth]{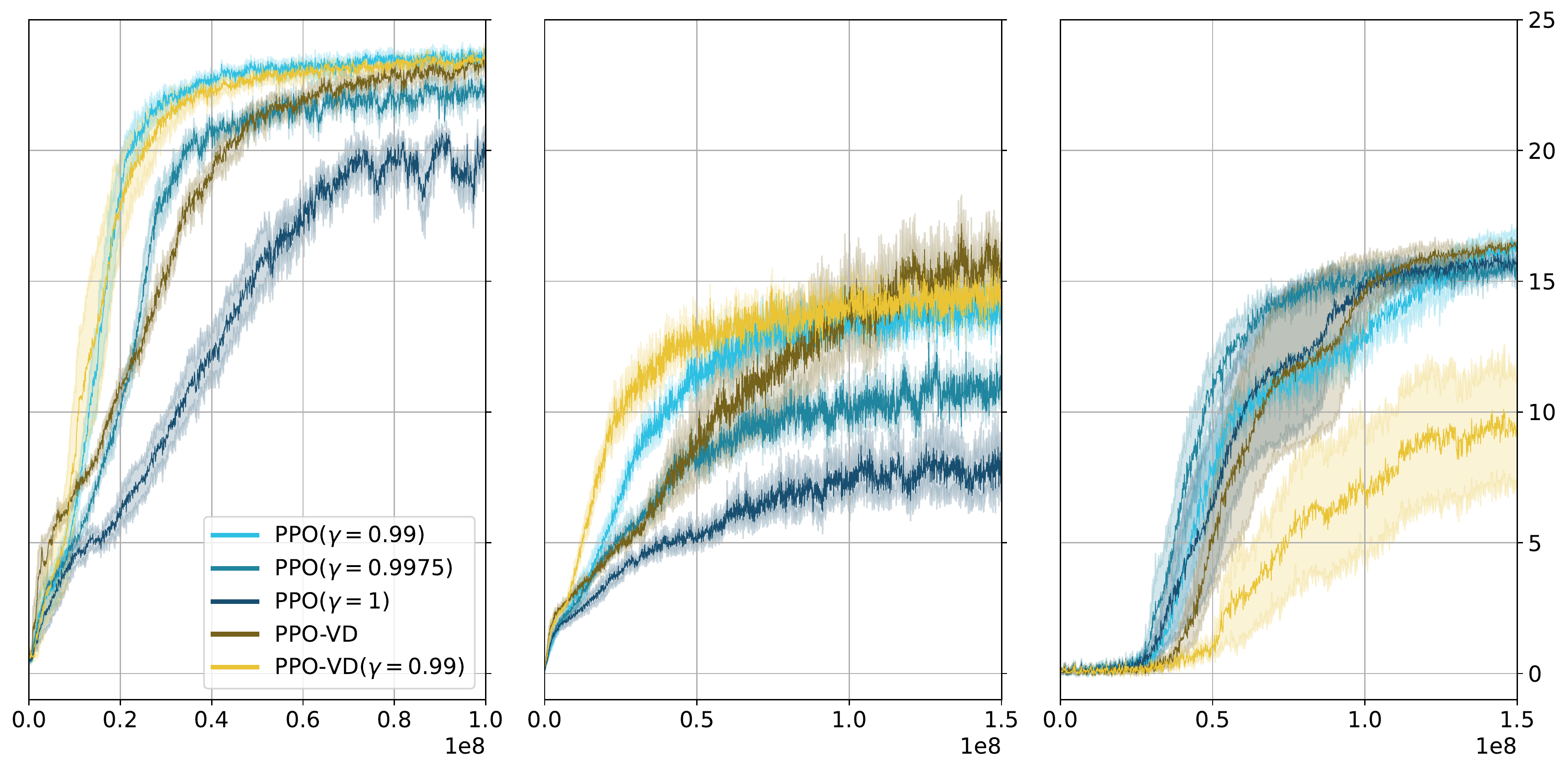}};
    
    \draw (-4.5,3.4) node {PointTSP}; 
    \draw (0,3.4) node {TimedTSP}; 
    \draw (4.5,3.4) node {ColourMatch}; 
    
    \draw (-6.8,0) node[rotate=90] {Return};
    \draw (0,-3.4) node {Frames};
    
    \end{tikzpicture}
    
    \caption{Learning curves for PPO and \vdn baselines. \vd{1} generally results in similar or better final performance compared to PPO, but \ppo{0.99} can be more sample efficient. Results are averaged over 5 runs, with error bars reporting standard error.}
    \label{fig:lcurves_ppo}

\end{figure}
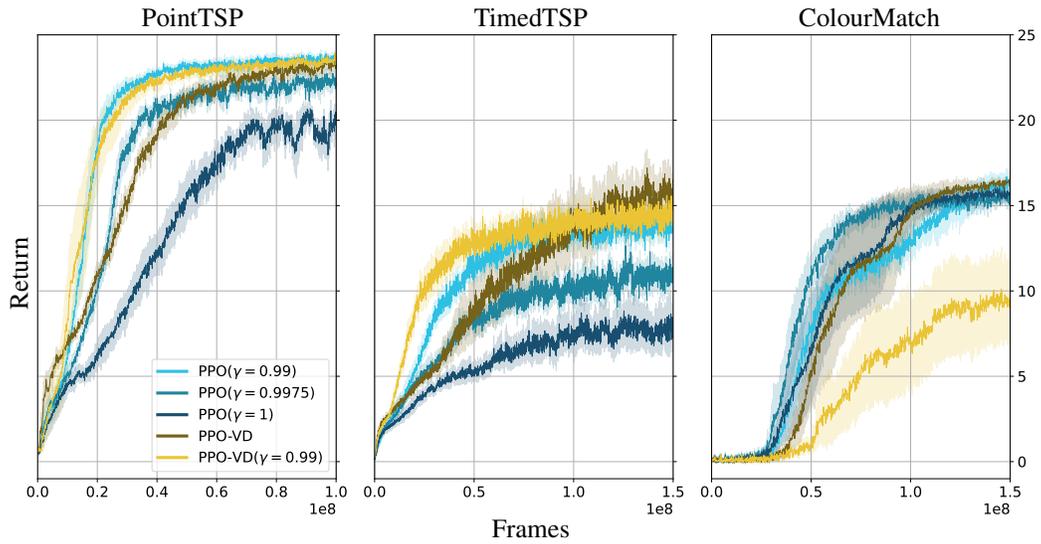

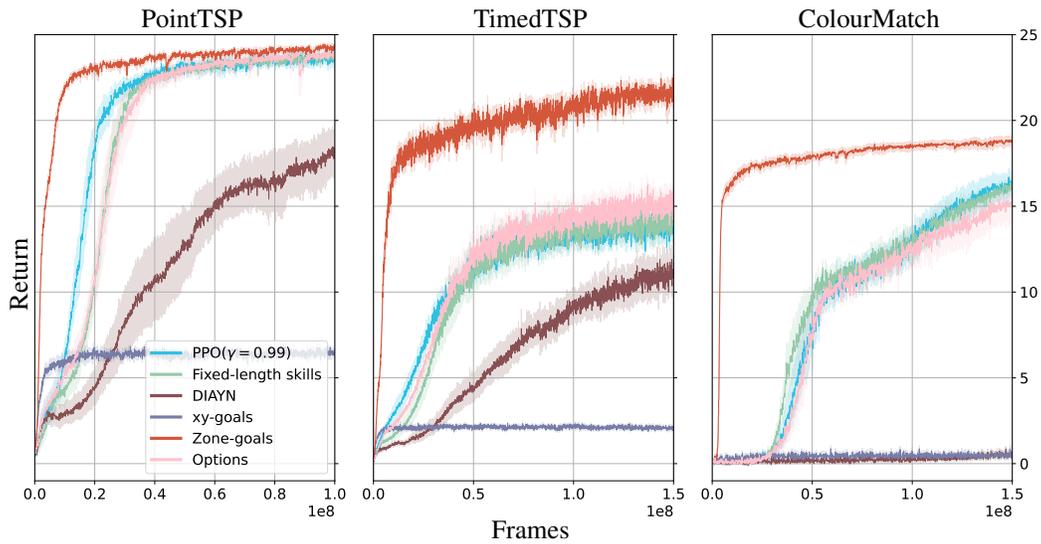
\begin{figure}
    \centering
    \begin{tikzpicture}
    \node at (0,0) {\includegraphics[width=0.98\columnwidth]{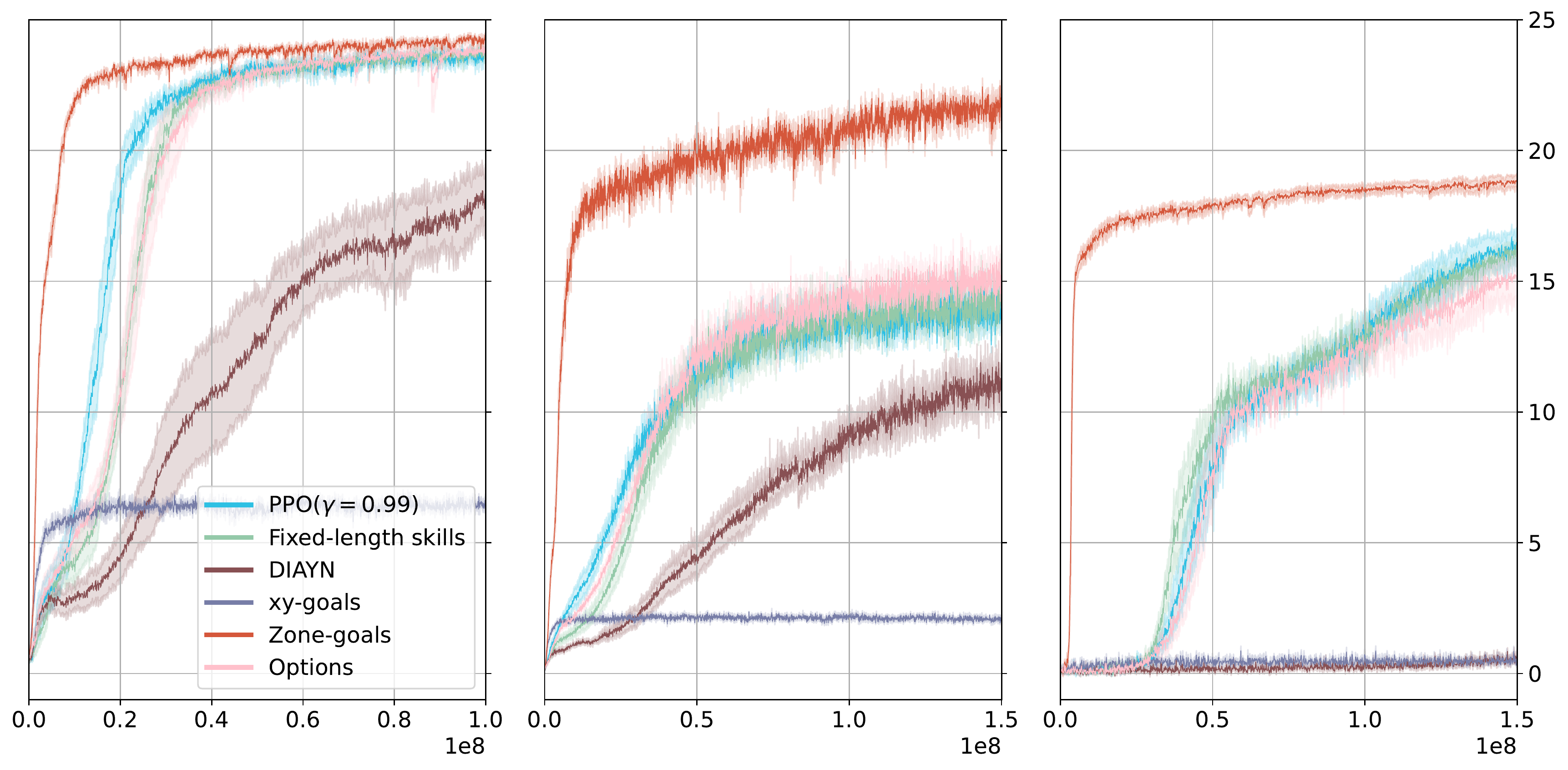}};
    
    \draw (-4.5,3.4) node {PointTSP}; 
    \draw (0,3.4) node {TimedTSP}; 
    \draw (4.5,3.4) node {ColourMatch}; 
    
    \draw (-6.8,0) node[rotate=90] {Return};
    \draw (0,-3.4) node {Frames};
    
    \end{tikzpicture}
    
    \caption{Learning curves for HRL baselines, with \ppo{0.99} included for comparison. Zone-goals significantly outperformed other baselines, while other HRL approaches performed similarly or worse compared to PPO. Results are averaged over 5 runs, with error bars reporting standard error.}
    \label{fig:lcurves_hrl}

\end{figure}

We show the learning curves of PPO and HRL baselines in Figures \ref{fig:lcurves_ppo} and \ref{fig:lcurves_hrl} respectively.

\subsection{Network Architectures}
Here, we report the network architectures for all methods in our experiments. The sizes of hidden layers are reported in the next section and were chosen to ensure the number of parameters of each method were roughly equal. All activations are ReLU.

\textbf{\ppo{0.99}, \ppo{0.9975}, \ppo{1}, \vd{0.99}, \vd{1}:} Observations were first encoded with the order-invariant neural architecture described in Section~\ref{subsection:architecture}: each zone is encoded using an MLP with two hidden layers, and then the average of all zone encodings is aggregated through a single linear layer. This observation encoding is then passed to the actor and critic (both single hidden-layer MLPs). For \vdn, the final critic layer has two output heads to model a mean and standard deviation (using softplus for the latter). 

\textbf{Fixed-length skills, DIAYN, Options:}
The high and low-level policies each use the same architecture as for PPO with minor changes: the hidden sizes are reduced to maintain the same number of parameters; the high-level policy outputs a discrete skill (5 possible values); and the low-level policy conditions on the current skill. For Options, the low-level policy additionally outputs a probability of terminating the current skill. For DIAYN, we additionally train a neural network to predict the current skill given next state. This network uses the order-invariant neural architecture: a two hidden-layer MLP for encoding zones and a one hidden-layer MLP for aggregation.

\textbf{xy-goals, Zone-goals:}
The high and low-level policies are the same as before with the following minor differences. In xy-goals, the high-level policy outputs xy-coordinates which the low-level policy conditions on. In Zone-goals, the high-level policy outputs a discrete zone, and the low-level policy conditions on the xy-coordinates of that zone.  

\subsection{Hyperparameters}

\newcolumntype{C}[1]{>{\centering\let\newline\\\arraybackslash\hspace{0pt}}m{#1}}

\begin{table}[]
\centering
\caption{Hyperparameters for PPO and \vdn for the majority of our experiments. The following alterations were made: on ColourMatch, the number of steps per update was doubled to 128000 (for all methods); on PointTSP, \vd{1} used 6 epochs for optimization. }
\label{table:ppo_hyper}
\begin{tabular}{@{}lccccc@{}}
\toprule
                       & \multicolumn{3}{c}{PPO} & \multicolumn{2}{c}{\vdn} \\ \cmidrule(l){2-4} \cmidrule(l){5-6}
Discount factor ($\gamma$)       & 0.99 &  0.9975 &  1 & 0.99 & 1 \\\midrule
Value loss coefficient & 0.5 & 0.5 & 0.5 & 0.005 & 0.005 \\
Env. steps per update  & \multicolumn{5}{c}{$\xleftarrow{\hspace*{0.05em}} 64,000 \xrightarrow{\hspace*{0.05em}}$} \\
Number of epochs       & \multicolumn{5}{c}{$\xleftarrow{\hspace*{0.05em}} 10 \xrightarrow{\hspace*{0.05em}}$}                             \\

Minibatch size         & \multicolumn{5}{c}{$\xleftarrow{\hspace*{0.05em}} 1,600 \xrightarrow{\hspace*{0.05em}}$}                             \\
Learning rate          & \multicolumn{5}{c}{$\xleftarrow{\hspace*{0.05em}} 3\times10^{-4} \xrightarrow{\hspace*{0.05em}}$}                             \\
GAE-$\lambda$          & \multicolumn{5}{c}{$\xleftarrow{\hspace*{0.05em}} 0.95 \xrightarrow{\hspace*{0.05em}}$}                             \\
Entropy coefficient    & \multicolumn{5}{c}{$\xleftarrow{\hspace*{0.05em}} 0.003 \xrightarrow{\hspace*{0.05em}}$}                             \\
Gradient Clipping      & \multicolumn{5}{c}{$\xleftarrow{\hspace*{0.05em}} 0.5 \xrightarrow{\hspace*{0.05em}}$}                             \\
PPO Clipping ($\varepsilon$) & \multicolumn{5}{c}{$\xleftarrow{\hspace*{0.05em}} 0.2 \xrightarrow{\hspace*{0.05em}}$}                             \\ \bottomrule
\end{tabular}
\end{table}

\newcolumntype{C}[1]{>{\centering\let\newline\\\arraybackslash\hspace{0pt}}m{#1}}

\begin{table}[t]
\small
\centering
\caption{Hyperparameters for HRL baselines (Fixed-length skills, DIAYN, Options, xy-goals, Zone-goals, TSP-solver). On ColourMatch, the number of steps per update was doubled to 128000 (for all methods). The high-level hyperparameters do not apply to TSP-solver, which did not have a high-level policy. }
\label{table:hrl_hyper}

\begin{tabular}{*{2}c}

\cmidrule[\heavyrulewidth]{1-2}

\multicolumn{1}{l}{Env. steps per update} & 64,000 \\
\\

\multicolumn{1}{l}{\textbf{(Low-level policy optimization)}} \\

\multicolumn{1}{l}{Number of epochs} & 10 \\

\multicolumn{1}{l}{Minibatch size} & 1,600 \\ 

\multicolumn{1}{l}{Discount factor ($\gamma$)} & 0.99 \\

\multicolumn{1}{l}{Learning rate} & $3\times10^{-4}$ \\ 

\multicolumn{1}{l}{GAE-$\lambda$} & 0.95 \\ 

\multicolumn{1}{l}{Entropy coefficient} & 0.003 \\

\multicolumn{1}{l}{Value loss coefficient} & 0.5 \\

\multicolumn{1}{l}{Gradient Clipping} & 0.5 \\

\multicolumn{1}{l}{PPO Clipping ($\varepsilon$)} & 0.1 \\

\multicolumn{1}{l}{                                                          }\\ 

\multicolumn{1}{l}{\textbf{(High-level policy optimization)}} \\

\multicolumn{1}{l}{Number of epochs} & 5 \\

\multicolumn{1}{l}{Minibatch size} & 80 \\

\multicolumn{1}{l}{Discount factor ($\gamma$)} & 1 \\

\multicolumn{1}{l}{Learning rate} & $3\times10^{-4}$ \\

\multicolumn{1}{l}{GAE-$\lambda$} & 0.95 \\

\multicolumn{1}{l}{Entropy coefficient} & 0.01 \\ 

\multicolumn{1}{l}{Value loss coefficient} & 0.5 \\

\multicolumn{1}{l}{Gradient Clipping} & 0.5 \\

\multicolumn{1}{l}{PPO Clipping ($\varepsilon$)} & 0.1 \\

\bottomrule
\end{tabular}
\end{table}

Hyperparameter tuning in RL is known to be laborious due to the large number of tunable hyperparameters, yet critical for performance. To ensure robust training, we set relatively large values for the number of env. steps per update and the minibatch size for all approaches while tuning the number of optimization epochs. The entropy coefficient was critical for final performance -- setting the value too high would indirectly reduce the agent's maximum speed while setting the value too low would lead to suboptimal convergence. Since many of our baselines are based on PPO, we found a similar set of hyperparameters to perform well across many methods.

We report the hyperparameters for PPO and \vdn in Table~\ref{table:ppo_hyper} and the hyperparameters for the HRL methods in Table~\ref{table:hrl_hyper}. For HRL methods with a fixed skill length (fixed-length skills, DIAYN, xy-goals), we chose a skill length of 200. For methods with a fixed number of discrete skills (fixed-length skills, DIAYN, Options) we set the number of skills to 5. 

DIAYN also used a hyperparameter $\alpha$ to linearly combine environment rewards with the diversity objective, which was set to $0.01$. We found that lower values of $\alpha$ caused the diversity objective to be ignored, resulting in similar performance to fixed-length skills.

\section{Broader Impact}

While our contributions are primarily foundational and may not directly lead to malicious use, we acknowledge that future applications that exploit our research could have the potential for a negative societal impact. In particular, the insights gathered from our results may help train reinforcement learning agents with greater capabilities, and therefore, greater potential for harm when misused. As many applications of reinforcement learning are safety-critical, including autonomous driving, robotics, and healthcare, we caution against over-trusting reinforcement learning, particularly in situations where the system was not trained on the environment in which it is being deployed. More work is needed to develop robust safety guarantees for RL before it can be applied in many real-world settings.

\end{document}